\pdfoutput=1
\documentclass[11pt]{article}
\usepackage[final]{acl}

\usepackage{times}
\usepackage{latexsym}
\usepackage[T1]{fontenc}
\usepackage[utf8]{inputenc}

\usepackage{amsfonts}
\usepackage{amsmath}
\usepackage{amssymb}        % enable \mathbb, \mathcal
\usepackage{bold-extra}     % enable \texttt{\textbf{}}
\usepackage{bm}             % enable bold font in math mode
\usepackage{graphicx}       % import graphical images
\usepackage{multirow}       % merge rows in tables
\usepackage{booktabs}       % borders in tabular
\usepackage{enumitem}       % list margins
\usepackage{subcaption}     % create sub-tables and sub-figures
\usepackage[hang,flushmargin]{footmisc}  % minimize footnote indentation

\usepackage{microtype}
\usepackage{hyperref}
\usepackage{url}
\usepackage{booktabs}
\usepackage{enumitem}
\usepackage{graphicx}
\usepackage{lineno}
\usepackage{xspace}
\usepackage{algorithm}
\usepackage{algpseudocode}
\usepackage{array}
\usepackage{longtable}
\usepackage{arydshln}
\usepackage{wrapfig}
\usepackage{caption}

\usepackage{natbib}
\usepackage{tcolorbox}

\usepackage{listings}
\usepackage{framed}
\usepackage{xcolor}
\usepackage{enumitem}

\definecolor{codegreen}{rgb}{0,0.6,0}
\definecolor{codegray}{rgb}{0.5,0.5,0.5}
\definecolor{codepurple}{rgb}{0.58,0,0.82}
\definecolor{backcolour}{rgb}{0.95,0.95,0.92}

\lstdefinestyle{jsonstyle}{
    backgroundcolor=\color{backcolour},
    commentstyle=\color{codegreen},
    keywordstyle=\color{magenta},
    numberstyle=\tiny\color{codegray},
    stringstyle=\color{codepurple},
    basicstyle=\ttfamily\footnotesize,
    breakatwhitespace=false,
    breaklines=true,
    captionpos=b,
    keepspaces=true,
    numbers=left,
    numbersep=5pt,
    showspaces=false,
    showstringspaces=false,
    showtabs=false,
    tabsize=2,
    morestring=[b]", % Strings are enclosed in "
    morekeywords={true,false,null}, % JSON keywords
}

\newtcolorbox{AIbox}[1][]{%
  colback=white,    % Background color
  colframe=black,   % Frame color
  boxrule=0.5pt,    % Frame thickness
  sharp corners,    % Sharp (non-rounded) corners
  #1                % Allow for additional options
}

\title{Search Wisely: Mitigating Sub-optimal Agentic Searches By Reducing Uncertainty}

{\author{Peilin Wu\textsuperscript{1}\thanks{Equal contribution}, Mian Zhang\textsuperscript{1}\footnotemark[1], Xinlu Zhang\textsuperscript{2}, Xinya Du\textsuperscript{1}, Zhiyu Zoey Chen\textsuperscript{1} \\
\textsuperscript{1}Department of Computer Science, The University of Texas at Dallas,\\
\textsuperscript{2}Department of Computer Science, University of California, Santa Barbara,\\
\texttt{\{peilin.wu, mian.zhang, zhiyu.chen2\}@utdallas.edu}
}}

\begin{document}
\maketitle

\begin{abstract}
Agentic Retrieval-Augmented Generation (RAG) systems enhance Large Language Models (LLMs) by enabling dynamic, multi-step reasoning and information retrieval. However, these systems often exhibit sub-optimal search behaviors like over-search (retrieving redundant information) and under-search (failing to initiate retrieval for necessary information), which hinder efficiency and reliability. This work formally defines and quantifies these behaviors, revealing their prevalence across multiple QA datasets and agentic RAG systems (e.g., one model could have avoided searching in 27.7\% of its search steps). Furthermore, we demonstrate a crucial link between these inefficiencies and the models' uncertainty regarding their own knowledge boundaries, where response accuracy correlates with model's uncertainty or confidence in its search decisions. To address this, we propose $\beta$-GRPO, a reinforcement learning-based training method that incorporates confidence threshold to reward high-certainty search decisions. Experiments on seven QA benchmarks show that $\beta$-GRPO enable a 3B model with better agentic RAG ability, outperforming other strong baselines with a 4\% higher average exact match score, with lower over-search and under-search rate\footnote{We have release our code for training at \url{https://github.com/mianzhang/Search-R1}.}.
% Our findings highlight the importance of knowledge boundary awareness in optimizing agentic RAG systems, offering a practical solution for enhancing their efficiency and accuracy.
\end{abstract}
\section{Introduction}
\label{sec:introduction}

Recent advances in Large Language Models (LLMs) have propelled their use in information-intensive tasks such as question answering and knowledge synthesis, especially when paired with retrieval capabilities \cite{wang2025chainofretrievalaugmentedgeneration}. Agentic Retrieval-Augmented Generation (RAG) frameworks~\cite{Jin2025-ev,Song2025-hj,Chen2025-nu} push this further by empowering LLMs to perform multi-step reasoning \cite{Li2025-ws} and dynamically decide when and what to retrieve \cite{guan2025deepragthinkingretrievalstep}, closely emulating sophisticated human research processes. However, despite these advancements, current agentic RAG systems often struggle with efficiency and reliability due to sub-optimal search behaviors \cite{shen-etal-2024-smartcal, qian2025smartselfawareagenttool, wang2025otcoptimaltoolcalls}. In particular, two major challenges: 1) over-search, where the model retrieves information it already knows , and 2) under-search, where it fails to seek external knowledge when necessary, have been identified as critical obstacles that degrade performance.

In this work, we conduct a thorough quantitative analysis to identify and measure the prevalence of over-search and under-search. Our experiments on several multi-hop QA datasets (2WikiMultiHopQA \cite{ho-etal-2020-constructing}, Bamboogle \cite{press-etal-2023-measuring}, HotpotQA \cite{yang-etal-2018-hotpotqa}, and MuSiQue \cite{trivedi-etal-2022-musique}) using contemporary LLMs like R1-Searcher \cite{Song2025-hj} and Search-R1 \cite{Jin2025-ev} reveal significant instances of sub-optimal search. We also further explore the connection between these behaviors and a model's awareness of its knowledge boundaries, finding that candidate responses generated with higher certainty about the necessity of a search query tend to achieve better accuracy.

To address this, we introduce $\beta$-GRPO, a variant of GRPO~\cite{Shao2024-dp} where the confidence of search calls are modeled as the minimal token probability of the search queries produced by the model and a confidence threshold is incorporated into the reward function, only encouraging generations with high-certainty search calls leading to correct answer. Through extensive experiments on seven QA benchmarks, we show that $\beta$-GRPO enables a 3B model with better agentic RAG ability compared to strong baselines with a 4\% higher average exact match score and 1.21\% fewer over-searches and 7.33\% fewer under-searches.
\section{Identifying Sub-optimal Search}
\label{sec:identifying-sub-optimal-search}
% \subsection{Empirical Analysis of Sub-optimal Search}
% \label{ssec:empirical-analysis-of-search-inefficiencies}
To investigate the prevalence of over-search and under-search, we conduct three experiments with the test sets of four widely recognized multihop QA datasets: 2WikiMultiHopQA \cite{ho-etal-2020-constructing}, Bamboogle \cite{press-etal-2023-measuring}, HotpotQA \cite{yang-etal-2018-hotpotqa}, and MuSiQue \cite{trivedi-etal-2022-musique}. We mainly investigate two recent LLMs that interact with search engines: R1-Searcher \cite{song2025r1searcherincentivizingsearchcapability} and Search-R1 \cite{jin2025searchr1trainingllmsreason}. We adopt the version trained based on Qwen2.5-7B \cite{qwen2025qwen25technicalreport} for a fair comparison.

\subsection{Formal Definition of Under-search \& Over-search}
\label{ssec:formal-definition-of-over-search-under-search}
Formally, let an LLM agent's interaction for a question be a sequence of steps $T = \{s_1, s_2, \ldots, s_N\}$. Each step $s_t$ comprises a reasoning component $r_t$. If the model decides to retrieve information, the retrieval step $s_t^R = (r_t, q_t, c_t)$ includes a search sub-query $q_t$ and the retrieved context $c_t = search(q_t)$. The sub-answer $a_t$ for this step $s_t^R$ is typically derived using $c_t$ and reflected in $r_{t+1}$. If the model does not retrieve, the non-retrieval step $s_t^{NR} = (r_t)$ relies on the existing context $\{s_1, s_2, \ldots, s_{t-1}\}$ and the model's internal knowledge $M$ to derive $a_t$ reflected in $r_{t}$. Let $a_t^*$ be the ground-truth answer step $s_t$. Over-search occurs if a retrieval step $s_t^R$'s answer $a_t$ could have been derived from $M$ and $\{s_1, s_2, \ldots, s_{t-1}\}$ only. Under-search occurs if a non-retrieval step $s_t^{NR}$ leads to $a_t \neq a_t^*$.

\subsection{Step-wise Analysis}
\label{ssec:step-wise-analysis}
\begin{figure}[t!]
\centering
\includegraphics[width=\columnwidth]{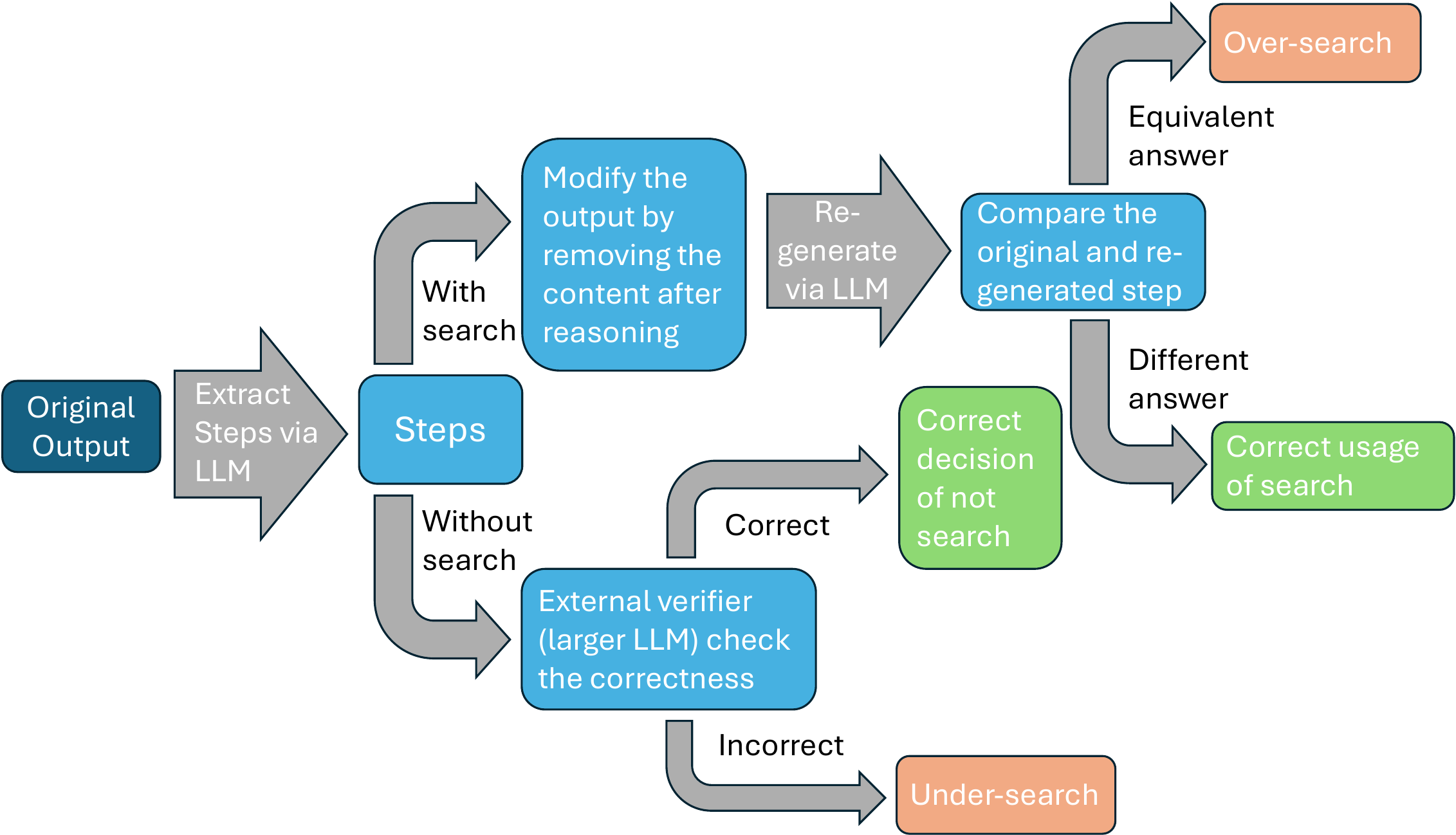}
\caption{Flowchart of analysis pipeline for over-search and under-search.}
\vspace{-0.1in}
\label{fig:pipeline}
\end{figure}

To directly measure whether a search step was truly necessary, we separate all outputs into individual steps and identify if each of them matches with the definition of over-search and under-search as described in Section \ref{ssec:formal-definition-of-over-search-under-search}. For over-search rate measurement, we prompted the model to answer sub-queries from all the steps with search behavior using only their internal knowledge and the preceding context. For under-search, we examine steps without searching and evaluate the correctness of the generated information. A detailed explanation of the analysis pipeline is provided in Appendix \ref{ssec:detailed-step-wise-analysis-procedure} with a flow chart in Figure \ref{fig:pipeline}. 

\paragraph{Capability to Answer from Memory}
% \label{ssec:capability-to-answer-from-memory}

\begin{figure}[t!]
\centering
\includegraphics[width=\columnwidth]{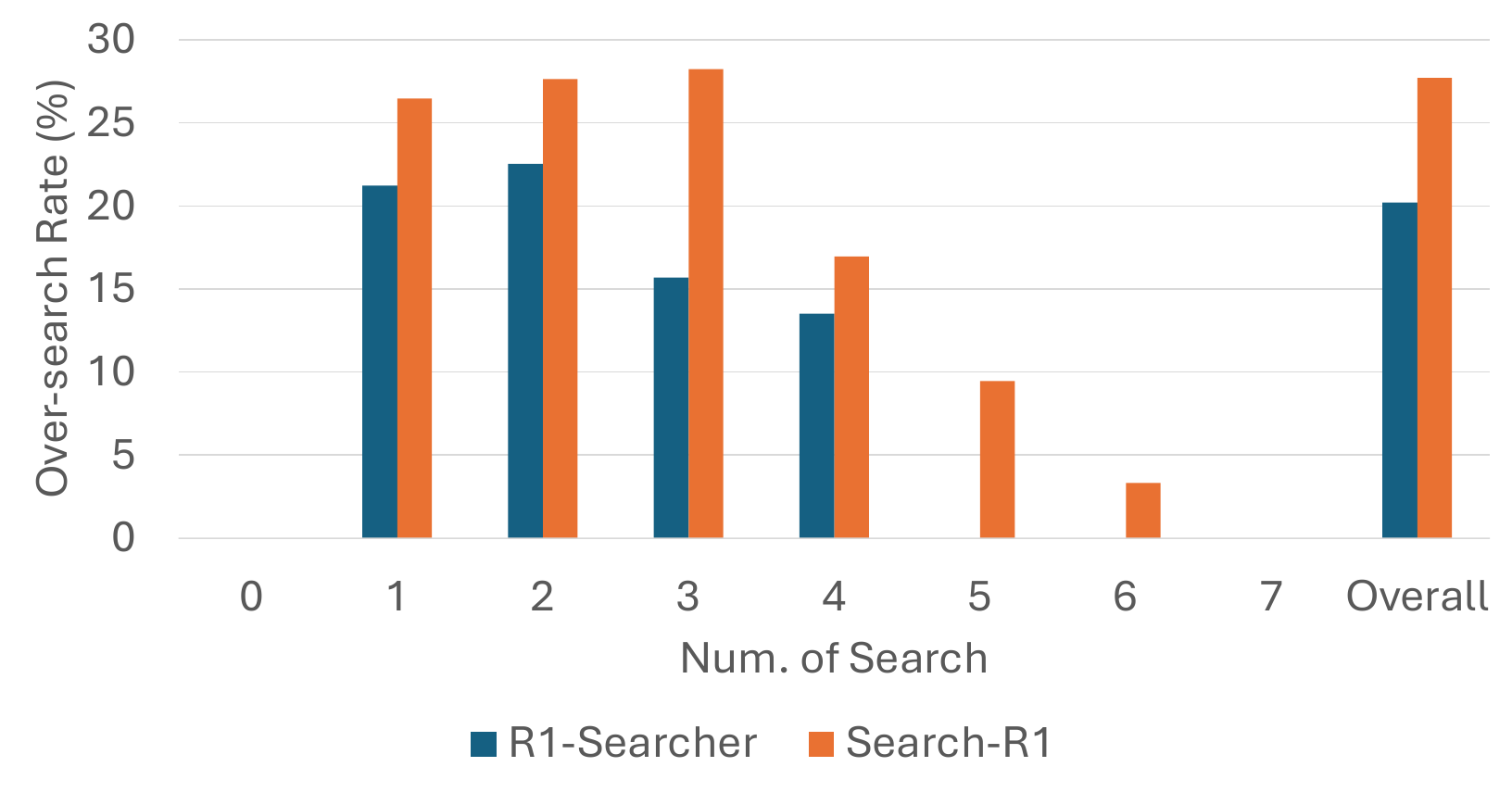}
\caption{Percentage for all search steps that can be answered without performing searches of R1-Searcher and Search-R1 on 4 datasets combined, with respect to the number of searches of each test sample.}
\vspace{-0.2in}
\label{fig:osr_search-r1_vs_r1-searcher}
\end{figure}

The results in Figure \ref{fig:osr_search-r1_vs_r1-searcher} show that a significant portion of search actions were instances of over-search. R1-Searcher could have answered correctly without searching in 20.2\% of its search steps overall, while Search-R1 could have done so in 27.7\% of its search steps. This highlights a substantial room for efficiency improvement. Figure \ref{fig:osr_search-r1_vs_r1-searcher} also shows the over-search rate for each subset of test samples grouped by the total number of search steps an agent used to solve an entire problem instance. 
% This analysis helps us understand if models that take more search steps overall are more or less prone to making unnecessary individual searches. For R1-Searcher, instances solved with 1-4 total searches all exhibit a notable percentage of over-search (ranging from 13.51\% to 22.53\%). This suggests that even when the model employs a seemingly reasonable number of total searches for a problem, a fair portion of those can be superfluous. Search-R1 shows a similar trend, with over-search rates remaining high (often above 26\% for problems solved in 1 to 3 total searches) and still present even in problems requiring more total search actions. 
The results per each subset indicates that over-search is a persistent issue irrespective of the overall search complexity adopted by the model for a given problem. 
Despite the step-wise analysis, we also conduct an analysis on comparing the number of searches versus the pre-given number of hops from the dataset in Appendix \ref{ssec:search-frequency-vs-optimal-hops}, which also supports our conclusion.

\paragraph{Error Rate in Non-Search Steps}
% \label{ssec:error-rate-in-non-search-steps}
\begin{figure}[t!]
\centering
\includegraphics[width=\columnwidth]{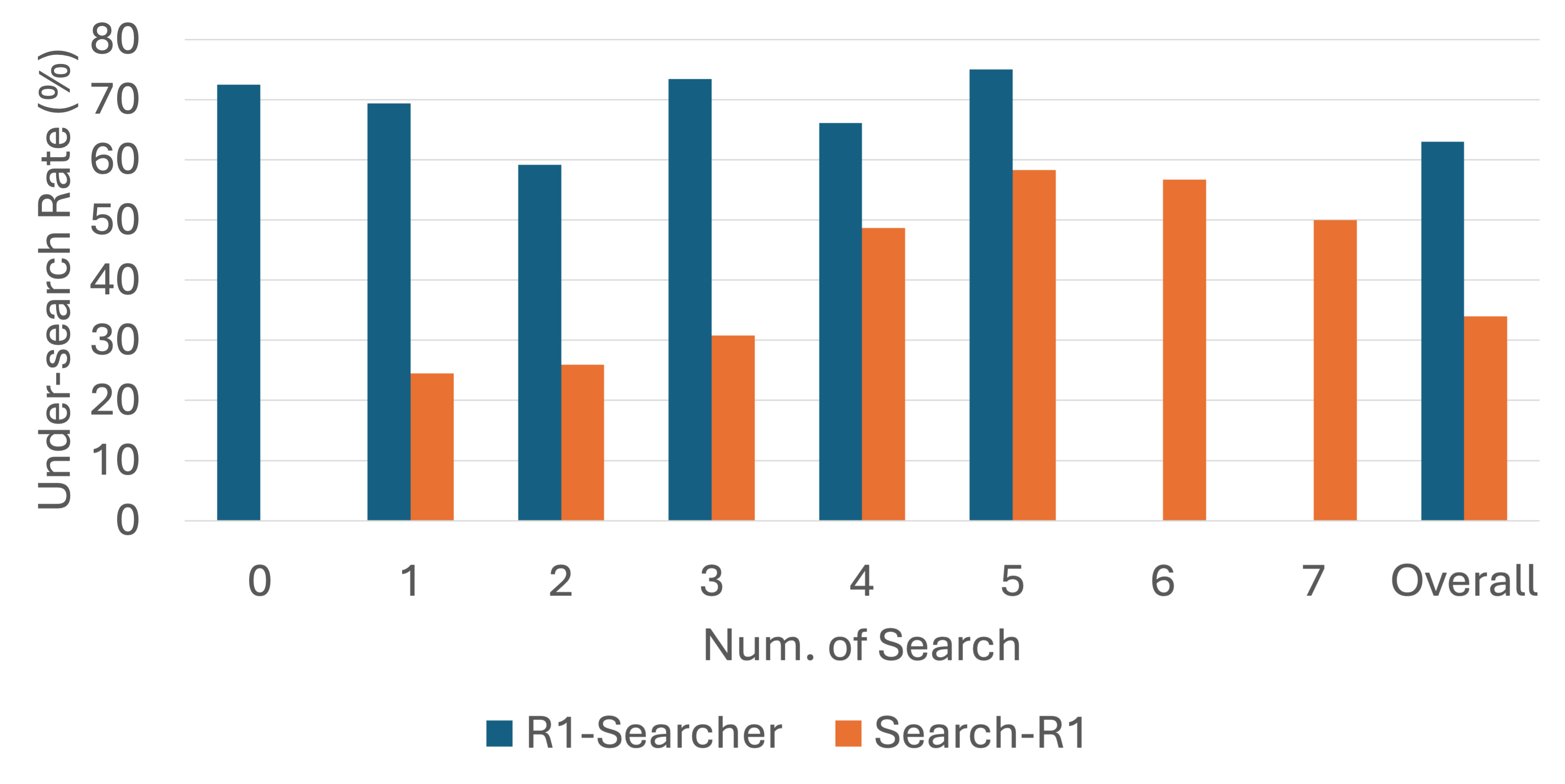}
\caption{Error rate for all non-search steps of R1-Searcher and Search-R1 on 4 datasets combined, with respect to the number of searches of each test sample.}
\vspace{-0.1in}
\label{fig:usr_search-r1_vs_r1-searcher}
\end{figure}

Figure \ref{fig:usr_search-r1_vs_r1-searcher} analyzes the error rate in non-search steps, which can be seen as the rate of under-search. 
Both models exhibited high error rates (R1-Searcher: 63\%, Search-R1: 33.98\%) in non-search steps, suggesting a strong tendency towards under-search leading to incorrect reasoning or hallucination. For R1-Searcher, this error rate was particularly high with fewer total searches (over 72\% if no searches were made). For Search-R1, errors in non-search steps remained notable even when performing many searches overall (e.g., 48.70\% for 4-search problems), possibly due to decision complexity in later stages. (See Figure \ref{fig:usr_search-r1_vs_r1-searcher} for detailed error rates by search step count).
%Both R1-Searcher and Search-R1 exhibits a very high error rate (63\% and 33.98\%) when it forgoes searching, suggesting a strong tendency towards under-search leading to incorrect internal reasoning or hallucination. For R1-Searcher with small number of total searches, the error rate in its non-search steps is extremely high (above 59\%, and over 72\% if no searches are made at all). This suggests that when R1-Searcher is too conservative with searching, it often does so at the cost of accuracy in the steps where it relies on internal knowledge. For Search-R1, which generally makes more searches, if a problem instance was resolved with only 1 total search, the non-search steps within that solution trace had a 24.49\% error rate. As the total number of searches Search-R1 uses for a problem increases, the error rate in its non-search steps also tends to increase, sometimes substantially (e.g., 48.70\% for 4-search problems, 58.33\% for 5-search problems). This might indicate that even when Search-R1 is performing many searches overall, the intermediate steps where it decides not to search are still prone to error, possibly because these non-search decisions occur in particularly complex situations or after multiple search iterations have already built up a confusing context.

\subsection{Sub-optimal Search \& Knowledge Boundary}
The observed tendencies towards over-search and under-search, combined with our definition, suggest a core deficiency in how agentic RAG models perceive knowledge boundaries—the limits of what they know versus what they need to find out. To illustrate the link between better knowledge boundary awareness and improved outcomes, we analyze the performance of 4 Qwen2.5-3B based Search-R1 models (including PPO and GRPO trained, Base and Instruct variants). We generate 5 candidate responses for each question and group these responses based on each output's \textbf{minimum probabilities within all the search query tokens in a trajectory} as the indication of certainty on knowledge boundary.
% Table \ref{tab:bon_confidence_accuracy} compares the final accuracy achieved by selecting the highest minimum logits from a pool of 5 responses characterized by lower average uncertainty versus a pool with lowest minimum logits (higher average uncertainty).

\begin{table}[t!]
\centering\resizebox{\columnwidth}{!}{
\begin{tabular}{ll|cccc}
\toprule
\bf Model Config & \bf Prob. Group & \bf 2Wiki & \bf Bamboogle & \bf HotpotQA & \bf Musique \\
\midrule
\multirow{2}{*}{Base + PPO} & Max & \bf{0.184} & 0.096 & \bf{0.152} & 0.038 \\
                          & Min  & 0.168 & 0.096 & 0.114 & 0.038 \\
\midrule
\multirow{2}{*}{Base + GRPO} & Max & \bf{0.249} & 0.112 & \bf{0.327} & \bf{0.085} \\
                           & Min  & 0.234 & 0.104 & 0.289 & 0.056 \\
\midrule
\multirow{2}{*}{Instruct + PPO} & Max & \bf{0.333}          & 0.250 & 0.262 & \bf{0.138} \\
                              & Min  & 0.297 & 0.250 & 0.262 & 0.116 \\
\midrule
\multirow{2}{*}{Instruct + GRPO} & Max & 0.402          & \bf{0.125}          & \bf{0.343}          & 0.116 \\
                               & Min  & 0.402          & 0.063 & 0.302 & 0.116 \\
\bottomrule
\end{tabular}}
\caption{Cover EM scores on multi-hop QA datasets, comparing groups of responses with higher vs. lower uncertainty (derived from average of minimum probability of search query tokens) on knowledge boundary. Bold indicates instances where the Max Prob. group achieved a strictly better performance.}
\vspace{-0.2in}
\label{tab:bon_confidence_accuracy}
\end{table}

As shown in Table \ref{tab:bon_confidence_accuracy}, candidate responses generated with lower intrinsic uncertainty generally lead to higher final accuracy (as high as 6\% on Bamboogle and 3.8\% on HotpotQA), across different training methods and base models. This suggests that when the model exhibits higher confidence (lower uncertainty) in its generation path, it is more likely to be on a correct trajectory. Therefore, improving an agent's ability to accurately gauge its internal knowledge state—effectively sharpening its knowledge boundary detection and reducing undue uncertainty—is a crucial step towards mitigating both over-search and under-search, thereby enhancing the overall efficiency and reliability of agentic RAG systems. Our approach is motivated by this principle, aiming to train agents to better assess and reduce uncertainty at each search decision.
\section{Approach}
\label{sec:approach}

\begin{table*}[t]
    \centering
    \scriptsize
    \renewcommand{\arraystretch}{1.2}
    \resizebox{0.9\textwidth}{!}{
        \begin{tabular}{lcccccccc}
            \toprule
            \textbf{Methods} & \multicolumn{3}{c}{\textbf{General QA}} & \multicolumn{4}{c}{\textbf{Multi-Hop QA}} \\
            \cmidrule{2-8}
             & \textbf{NQ$^\dagger$} & \textbf{TriviaQA$^\star$} & \textbf{PopQA$^\star$} & \textbf{HotpotQA$^\dagger$} & \textbf{2wiki$^\star$} & \textbf{Musique$^\star$} & \textbf{Bamboogle$^\star$} & \textbf{Average} \\
            \midrule
            Direct Prompting & 0.106 & 0.288 & 0.108 & 0.149 & 0.244 & 0.020 & 0.024 & 0.134 \\
            CoT Prompting & 0.023 & 0.032 & 0.005 & 0.021 & 0.021 & 0.002 & 0.000 & 0.015 \\
            IRCoT & 0.111 & 0.312 & 0.200 & 0.164 & 0.171 & 0.067 & 0.240 & 0.181 \\
            Search-o1 & 0.238 & 0.472 & 0.262 & 0.221 & 0.218 & 0.054 & \textbf{0.320} & 0.255 \\
            RAG & 0.348 & 0.544 & 0.387 & 0.255 & 0.226 & 0.047 & 0.080 & 0.270  \\
            SFT & 0.249 & 0.292 & 0.104 & 0.186 & 0.248 & 0.044 & 0.112 & 0.176  \\
            % \hdashline
            % \multicolumn{8}{l}{\textbf{Qwen2.5-3b-Base}} \\
            R1 & 0.226 & 0.455 & 0.173 & 0.201 & 0.268 & 0.055 & 0.224 & 0.229  \\
            Search-R1 & 0.406 & 0.587 & 0.435 & 0.284 & 0.273 & 0.049 & 0.088 & 0.303  \\
            \hline
            Search-R1-GRPO & 0.432 & 0.578 & 0.413 & 0.294 & 0.271 & 0.067 & 0.112 & 0.309  \\
            Search-R1-$\beta$-GRPO (ours) & \textbf{0.468} & \textbf{0.625} & \textbf{0.449} & \textbf{0.334} & \textbf{0.304} & \textbf{0.086} & 0.144 & \textbf{0.344}  \\
            % \hline
            % \multicolumn{8}{l}{\textbf{Qwen2.5-3b-Instruct}} \\
            % R1-instruct & 0.210 & 0.449 & 0.171 & 0.208 & 0.275 & 0.060 & 0.192 & 0.224  \\
            % Search-R1-instruct & 0.341 & 0.545 & 0.378 & 0.324 & \textbf{0.319} & \textbf{0.103} & \textbf{0.264} & 0.325  \\
            % Search-R1-instruct-C &  &  &  &  &  &  &  &  \\
            % Search-R1-instruct-CU &  &  &  &  &  &  &  &  \\
            \bottomrule
        \end{tabular}
    }
    \caption{Main results. The best performance is set in bold. $^\dagger/^\star$ represents in-domain/out-domain datasets.}\label{tab:main}
\end{table*}

Current RL powered agentic RAG methods~\cite{Jin2025-ev,Song2025-hj,Chen2025-nu} do not explicitly model the knowledge self-awareness during the training process, resulting in generations with low confidence, which are not desired and shown to easily contain wrong answer compared to generations with higher confidence (Table \ref{tab:bon_confidence_accuracy}). To this end, we propose a simple yet effective variant of GRPO~\cite{Shao2024-dp}, $\beta$-GRPO, which leverages the uncertainty of the search query spans for more effective rewarding and training.

\noindent\textbf{Agentic RAG with RL (Search-R1~\cite{Jin2025-ev})} Given a question, we prompt the policy model to explicitly reason enclosed within \verb|<think></think>| tags about whether to use an off-the-shelf search tool, and, if so, to generate a search query within \verb|<search></search>| tags. The search tool then returns relevant documents inside \verb|<information></information>| tags. Once obtaining new information, the policy model can either continue searching for additional information or provide a final answer within \verb|<answer></answer>| tags. The instruction given to the policy model could be found in Appendix~\ref{apx:instruction}. If the final answer match the groundtruth, the response will be given a reward 1, otherwise 0. And the policy are updated via policy gradient methods like GRPO~\cite{Shao2024-dp}.

\noindent\textbf{$\beta$-GRPO} Motivated by the observation that rollouts with low-confidence search calls are more likely to be incorrect, we incorporate model confidence into the RL reward process. Specifically, for each rollout containing search calls (enclosed within \verb|<search></search>| tags), we extract the probabilities of the search tokens including the tags and use the minimum probability among them as a measure of the model confidence for the search calls within a rollout~\cite{Jiang2023-mk}. We then set a confidence threshold $\beta$: only rollouts with the confidence of search calls (if exist) above $\beta$ and correct answers receive a reward of 1, otherwise 0. Formally, for a given reasoning trajectory $T={s_1, s_2, ..., s_N}$ as described in Section \ref{ssec:formal-definition-of-over-search-under-search}, with predicted final answer $a_f$, ground-truth final answer $a^*_f$, and confidence $C(T)$  the reward $R(T)$ is calculated as Equation \ref{eq:reward},

\begin{equation}
    R(T)=
    \begin{cases}
        1, & \text{if $a_f = a^*_f \wedge C(T) > \beta$} \\
        0, & \text{otherwise}
    \end{cases}
\label{eq:reward}
\end{equation}

where $C(T)$ is the confidence of trajectory $T$ calculated as Equation \ref{eq:confidence},

\begin{equation}
C(T) = \min_{s_t^R \in T, w \in q_t} P(w)
\label{eq:confidence}
\end{equation}

where $w$ is the token that makes up the search query $q_t$ for a given retrieval step and $P(w)$ is the probability assigned to $w$.

\section{Experiments}\label{sec:experiments}

\noindent\textbf{Datasets} We follow Search-R1~\cite{Jin2025-ev} using a mixture of the NQ~\cite{Kwiatkowski2019-zk} and HotpotQA~\cite{yang-etal-2018-hotpotqa} training sets for model training. For evaluation, we consider seven QA benchmarks, including general QA datasets, NQ, TriviaQA~\cite{Joshi2017-qg}, and PopQA~\cite{Mallen2023-eq}, as well as multi-hop QA datasets: HotpotQA, 2WikiMultiHopQA~\cite{ho-etal-2020-constructing}, Bamboogle~\cite{press-etal-2023-measuring}, and MuSiQue~\cite{trivedi-etal-2022-musique}. Exact match (EM) is used as our main evaluation metric.

% \begingroup
% \setlength{\tabcolsep}{4pt}
% \begin{table*}[t]
% \scriptsize
% \renewcommand{\arraystretch}{1.2} % Increase the row height by 2 times
% \begin{center}
% \resizebox{0.95\textwidth}{!}{
% \begin{tabular}{p{7cm}p{7cm}}
% \toprule
% Search-R1-GRPO & Search-R1-$\beta$-GRPO (ours) \\
% \midrule
% \textbf{Question}: In what Country is Sul America Esporte Clube in?
% \textbf{Generation}: <think> The question asks for the country hosting Sul America Esporte Clube, a team in the brazilian C1 and C2 league, thus the direct reasoning is likely to be Brazilian football team residing in Brazil. 
% &
% \textbf{Question}: In what Country is Sul America Esporte Clube in?
% \textbf{Generation}: <search> in what Country is Sul America Esporte Clube in </search>
% <information>
% Doc 1: ...
% Doc 2: ...
% Doc 3: ...
% </information>
% <answer> Brazil </answer> 
% (Confidence: 0.99) \\

% \bottomrule
% \end{tabular}
% }
% \end{center}
% \caption{Case Study. The retrieved documents are omitted for clarity.}\label{tab:case}
% \end{table*}
% \endgroup

\begingroup
\setlength{\tabcolsep}{4pt}
\begin{table*}[t]
\scriptsize
\renewcommand{\arraystretch}{1.2} % Increase the row height
\begin{center}
\resizebox{0.95\textwidth}{!}{
\begin{tabular}{p{7cm}p{7cm}}
\toprule
\textbf{Baseline Model (Search-R1-GRPO)} & \textbf{Our Model (Search-R1-$\beta$-GRPO)} \\
\midrule
\multicolumn{2}{c}{\textit{\textbf{Case 1: Mitigating Over-search for common knowledge questions}}} \\
\midrule
\textbf{Question}: Who was the first president of the United States? & \textbf{Question}: Who was the first president of the United States? \\
\textbf{Generation}: <think> The user is asking a question about a historical figure. To ensure accuracy, I will perform a search. </think> <search> first president of the United States </search> & \textbf{Generation}: <think> The user is asking for the first president of the United States. This is a well-established fact stored in my parametric knowledge. I am highly confident in the answer and do not need to perform an external search. </think> \\<information>Doc 1: ... Doc 2: ... </information> & <answer> George Washington </answer> \\
<answer> George Washington </answer> & \\
\textbf{Correctness}: Correct & \textbf{Correctness}: Correct \\
\midrule
\multicolumn{2}{c}{\textit{\textbf{Case 2: Mitigating Under-search for knowledge-intensive questions}}} \\
\midrule
\textbf{Question}: In what Country is Sul America Esporte Clube in? & \textbf{Question}: In what Country is Sul America Esporte Clube in? \\
\textbf{Generation}: <think> The question asks for the country of Sul America Esporte Clube. The name "Sul America" sounds Portuguese or Spanish, suggesting a location in South America. It could be in Argentina or Colombia. I will make a guess based on common football leagues. </think> & \textbf{Generation}: <think> The user is asking for the location of a specific sports club, "Sul America Esporte Clube". This is a niche entity, and my internal knowledge is likely insufficient. To provide an accurate answer, I must perform a search. </think> <search> country of Sul America Esporte Clube </search> \\
<answer> Argentina </answer> & <information> Doc 1: ... Doc 2: ... </information> \\
& <answer> Brazil </answer> \\
& (Confidence: 0.99) \\
\textbf{Correctness}: Incorrect & \textbf{Correctness}: Correct \\
\bottomrule
\end{tabular}
}
\end{center}
\caption{Case Study comparing search behaviors. The top pair demonstrates how our model avoids an unnecessary search (over-search) for a common fact. The bottom pair shows our model correcting a failure to search (under-search), preventing hallucination for a niche query. Retrieved documents are omitted or summarized for clarity.}\label{tab:case}
\end{table*}
\endgroup

% \zhiyu{Emphasize this is an under-search example and explain how our method improve it by issue proper search. }
\noindent\textbf{Baselines} We compare our method with several baselines: methods that do not use a retriever including direct prompting, Chain-of-Thought (CoT)~\cite{Wei2022-ds} prompting, supervised fine-tuning (SFT)~\cite{Chung2022-ls}, and reinforcement learning-based fine-tuning (R1)~\cite{DeepSeek-AI2025-te}; methods that use a retriever but do not perform agentic retrieval, such as Retrieval-Augmented Generation (RAG)~\cite{Lewis2020-ik} and IRCoT~\cite{Trivedi2023-cl}; and finally, agentic retrieval methods, including Search-o1~\cite{Li2025-ws} and Search-R1~\cite{Jin2025-ev}.

Based on our preliminary experiments, we found that training the policy model from scratch using our confidence-based rewards prevents it from learning effective search behavior. Therefore, we use Qwen2.5-3B~\cite{qwen2025qwen25technicalreport} and initialize it with the parameters from Search-R1. Then we continue training using GRPO with different reward functions: one using the original answer-based reward (Search-R1-GRPO), and the other using our proposed confidence-based reward (Search-R1-$\beta$-GRPO). We set the value of $\beta$ as 0.4 according to the analysis in Section~\ref{sec:analysis}. Detailed training configurations could be found in Appendix~\ref{apx:train_config}.

\noindent\textbf{Results} As shown in Table~\ref{tab:main}, agentic search with RL training (Search-R1*) significantly outperforms other baselines, indicating that incorporating search through autonomous reasoning and RL training is more effective than non-agentic or prompting methods. Our model, Search-R1-$\beta$-GRPO, achieves the highest overall average EM score across the datasets. Figure~\ref{fig:reward_curve} in Appendix \ref{apx:train_config} shows the training rewards for Search-R1-GRPO and Search-R1-$\beta$-GRPO. We observe that the rewards for Search-R1-GRPO fluctuate and do not show clear improvement over training steps. In contrast, Search-R1-$\beta$-GRPO achieves higher and more stable rewards. This improved performance suggests that our proposed reward assignment based on the confidence of search calls within a rollout is effective.

\section{Analysis}\label{sec:analysis}
% \zhiyu{You can add a summary for these two paragraphs, highlight at the beginning.}
\noindent\textbf{Ablation on $\beta$ \& Case Study} Following~\citet{Jiang2023-mk}, we experiment with three confidence threshold values: 0.2, 0.4, and 0.6. The average EM scores are 0.341, 0.344 and 0.336 with a threshold of 0.4 yields the best result. Moreover, we find 115 test cases from the multi-hop QA datasets where Search-R1-$\beta$-GRPO produces a correct answer with higher confidence, while Search-R1-GRPO gives an incorrect answer. These cases clearly benefit from the increased model confidence enabled by the proposed $\beta$-GRPO. 

% An example is shown in Table~\ref{tab:case}: Search-R1-GRPO lacks confidence and fails to provide a definite answer, whereas Search-R1-$\beta$-GRPO generates a confident search query and produces the correct answer.

\noindent\textbf{Under-searches \& Over-searches} We also measure the rate of over-search and under-search of our Search-R1-$\beta$-GRPO and the baseline Search-R1-GRPO trained based on Qwen2.5-3B with the methods in Section \ref{ssec:step-wise-analysis}. Compared with Search-R1-GRPO, which has overall 21.10\% over-search rate and 42.04\% under-search rate\%, our Search-R1-$\beta$-GRPO achieves 19.89\% over-search rate and 34.71\% under-search rate, which are lower than the baseline method. This shows that our method effectively reduces both types of sub-optimal searches. 
% \zhiyu{Add more case studies for both over and under search in appendix. You can also draw detailed search number break downs like figure 1 and 2 in appendix. }

\noindent\textbf{Case Study}
Our case study in Table \ref{tab:case} highlights the model's improved search decisions. For a simple question ("Who was the first president?"), the baseline model performs an unnecessary search, whereas our model confidently answers from its internal knowledge. Conversely, when faced with an obscure query ("In what Country is Sul America Esporte Clube in?"), the baseline hallucinates an incorrect answer. Our model correctly identifies this knowledge gap, initiates a search, and provides the accurate answer.
\section{Conclusion}
\label{sec:conclusion}
% In this work, we formally define and quantify sub-optimal search behaviors, over-search and under-search, in agentic RAG systems, revealing their prevalence and impact. By introducing $\beta$-GRPO, a confidence-aware policy gradient method, we enable a 3B model with better agentic RAG ability than strong baselines.
In this work, we formally define sub-optimal search behaviors, over-search and under-search, in agentic RAG systems. Our analysis showed these behaviors are widespread; for instance, one model could have avoided searching in 27.7\% of its search steps, while another exhibited error rates as high as 63\% in non-search steps, indicating significant under-searching. We established a link between these inefficiencies and a model's uncertainty about its knowledge boundaries, finding that higher confidence in search decisions correlates with better accuracy. By introducing $\beta$-GRPO, a confidence-aware policy gradient method, we enable a 3B model with better agentic RAG ability than strong baselines. This approach, which rewards only high-certainty search decisions that lead to correct answers, resulted in a 4\% higher average exact match score and notable reductions in both over-search and under-search rates. Future work should explore more sophisticated and fine-grained reward design on trajectory, with experiments on larger size models. 
\section*{Limitations}
We formally define and quantify sub-optimal search behaviors in agentic RAG systems and propose $\beta$-GRPO to train agentic RAG models with improved self-knowledge awareness. However, we acknowledge that sub-optimal search behaviors, over-search and under-search, are persistent challenges that require further investigation, especially in more open-ended tasks like deep research~\cite{Alzubi2025-wj}. Additionally, due to limited computational resources, we are unable to train larger models and leave it for future work.
\section*{Acknowledgements}
We thank the anonymous reviewers for their thoughtful comments and suggestions. We are also grateful to our colleagues and collaborators for their insightful discussions and support throughout this project.
% To be filled.

% References come after the acknowledgment section
\bibliography{custom,paperpile}

\begin{thebibliography}{30}
\expandafter\ifx\csname natexlab\endcsname\relax\def\natexlab#1{#1}\fi

\bibitem[{Alzubi et~al.(2025)Alzubi, Brooks, Chiniya, Contente, von Gerlach, Irwin, Jiang, Kaz, Nguyen, Oh, Tyagi, and Viswanath}]{Alzubi2025-wj}
Salaheddin Alzubi, Creston Brooks, Purva Chiniya, Edoardo Contente, Chiara von Gerlach, Lucas Irwin, Yihan Jiang, Arda Kaz, Windsor Nguyen, Sewoong Oh, Himanshu Tyagi, and Pramod Viswanath. 2025.
\newblock Open deep search: Democratizing search with open-source reasoning agents.
\newblock \emph{arXiv [cs.LG]}.

\bibitem[{Chen et~al.(2025)Chen, Li, Sun, Zhou, Zhu, Yang, Zhou, Chen, Wang, Pan, Zhang, and Chen}]{Chen2025-nu}
Mingyang Chen, Tianpeng Li, Haoze Sun, Yijie Zhou, Chenzheng Zhu, Fan Yang, Zenan Zhou, Weipeng Chen, Haofen Wang, Jeff~Z Pan, Wen Zhang, and Huajun Chen. 2025.
\newblock {ReSearch}: Learning to reason with search for {LLMs} via reinforcement learning.
\newblock \emph{arXiv [cs.AI]}.

\bibitem[{Chung et~al.(2022)Chung, Hou, Longpre, Zoph, Tay, Fedus, Li, Wang, Dehghani, Brahma, Webson, Gu, Dai, Suzgun, Chen, Chowdhery, Castro-Ros, Pellat, Robinson, Valter, Narang, Mishra, Yu, Zhao, Huang, Dai, Yu, Petrov, Chi, Dean, Devlin, Roberts, Zhou, Le, and Wei}]{Chung2022-ls}
Hyung~Won Chung, Le~Hou, Shayne Longpre, Barret Zoph, Yi~Tay, William Fedus, Yunxuan Li, Xuezhi Wang, Mostafa Dehghani, Siddhartha Brahma, Albert Webson, Shixiang~Shane Gu, Zhuyun Dai, Mirac Suzgun, Xinyun Chen, Aakanksha Chowdhery, Alex Castro-Ros, Marie Pellat, Kevin Robinson, Dasha Valter, Sharan Narang, Gaurav Mishra, Adams Yu, Vincent Zhao, Yanping Huang, Andrew Dai, Hongkun Yu, Slav Petrov, Ed~H Chi, Jeff Dean, Jacob Devlin, Adam Roberts, Denny Zhou, Quoc~V Le, and Jason Wei. 2022.
\newblock Scaling instruction-finetuned language models.
\newblock \emph{arXiv [cs.LG]}.

\bibitem[{{DeepSeek-AI} et~al.(2025){DeepSeek-AI}, Guo, Yang, Zhang, Song, Zhang, Xu, Zhu, Ma, Wang, Bi, Zhang, Yu, Wu, Wu, Gou, Shao, Li, Gao, Liu, Xue, Wang, Wu, Feng, Lu, Zhao, Deng, Zhang, Ruan, Dai, Chen, Ji, Li, Lin, Dai, Luo, Hao, Chen, Li, Zhang, Bao, Xu, Wang, Ding, Xin, Gao, Qu, Li, Guo, Li, Wang, Chen, Yuan, Qiu, Li, Cai, Ni, Liang, Chen, Dong, Hu, Gao, Guan, Huang, Yu, Wang, Zhang, Zhao, Wang, Zhang, Xu, Xia, Zhang, Zhang, Tang, Li, Wang, Li, Tian, Huang, Zhang, Wang, Chen, Du, Ge, Zhang, Pan, Wang, Chen, Jin, Chen, Lu, Zhou, Chen, Ye, Wang, Yu, Zhou, Pan, Li, Zhou, Wu, Ye, Yun, Pei, Sun, Wang, Zeng, Zhao, Liu, Liang, Gao, Yu, Zhang, Xiao, An, Liu, Wang, Chen, Nie, Cheng, Liu, Xie, Liu, Yang, Li, Su, Lin, Li, Jin, Shen, Chen, Sun, Wang, Song, Zhou, Wang, Shan, Li, Wang, Wei, Zhang, Xu, Li, Zhao, Sun, Wang, Yu, Zhang, Shi, Xiong, He, Piao, Wang, Tan, Ma, Liu, Guo, Ou, Wang, Gong, Zou, He, Xiong, Luo, You, Liu, Zhou, Zhu, Xu, Huang, Li, Zheng, Zhu, Ma, Tang, Zha, Yan, Ren, Ren, Sha, Fu, Xu, Xie,
  Zhang, Hao, Ma, Yan, Wu, Gu, Zhu, Liu, Li, Xie, Song, Pan, Huang, Xu, Zhang, and Zhang}]{DeepSeek-AI2025-te}
{DeepSeek-AI}, Daya Guo, Dejian Yang, Haowei Zhang, Junxiao Song, Ruoyu Zhang, Runxin Xu, Qihao Zhu, Shirong Ma, Peiyi Wang, Xiao Bi, Xiaokang Zhang, Xingkai Yu, Yu~Wu, Z~F Wu, Zhibin Gou, Zhihong Shao, Zhuoshu Li, Ziyi Gao, Aixin Liu, Bing Xue, Bingxuan Wang, Bochao Wu, Bei Feng, Chengda Lu, Chenggang Zhao, Chengqi Deng, Chenyu Zhang, Chong Ruan, Damai Dai, Deli Chen, Dongjie Ji, Erhang Li, Fangyun Lin, Fucong Dai, Fuli Luo, Guangbo Hao, Guanting Chen, Guowei Li, H~Zhang, Han Bao, Hanwei Xu, Haocheng Wang, Honghui Ding, Huajian Xin, Huazuo Gao, Hui Qu, Hui Li, Jianzhong Guo, Jiashi Li, Jiawei Wang, Jingchang Chen, Jingyang Yuan, Junjie Qiu, Junlong Li, J~L Cai, Jiaqi Ni, Jian Liang, Jin Chen, Kai Dong, Kai Hu, Kaige Gao, Kang Guan, Kexin Huang, Kuai Yu, Lean Wang, Lecong Zhang, Liang Zhao, Litong Wang, Liyue Zhang, Lei Xu, Leyi Xia, Mingchuan Zhang, Minghua Zhang, Minghui Tang, Meng Li, Miaojun Wang, Mingming Li, Ning Tian, Panpan Huang, Peng Zhang, Qiancheng Wang, Qinyu Chen, Qiushi Du, Ruiqi Ge, Ruisong
  Zhang, Ruizhe Pan, Runji Wang, R~J Chen, R~L Jin, Ruyi Chen, Shanghao Lu, Shangyan Zhou, Shanhuang Chen, Shengfeng Ye, Shiyu Wang, Shuiping Yu, Shunfeng Zhou, Shuting Pan, S~S Li, Shuang Zhou, Shaoqing Wu, Shengfeng Ye, Tao Yun, Tian Pei, Tianyu Sun, T~Wang, Wangding Zeng, Wanjia Zhao, Wen Liu, Wenfeng Liang, Wenjun Gao, Wenqin Yu, Wentao Zhang, W~L Xiao, Wei An, Xiaodong Liu, Xiaohan Wang, Xiaokang Chen, Xiaotao Nie, Xin Cheng, Xin Liu, Xin Xie, Xingchao Liu, Xinyu Yang, Xinyuan Li, Xuecheng Su, Xuheng Lin, X~Q Li, Xiangyue Jin, Xiaojin Shen, Xiaosha Chen, Xiaowen Sun, Xiaoxiang Wang, Xinnan Song, Xinyi Zhou, Xianzu Wang, Xinxia Shan, Y~K Li, Y~Q Wang, Y~X Wei, Yang Zhang, Yanhong Xu, Yao Li, Yao Zhao, Yaofeng Sun, Yaohui Wang, Yi~Yu, Yichao Zhang, Yifan Shi, Yiliang Xiong, Ying He, Yishi Piao, Yisong Wang, Yixuan Tan, Yiyang Ma, Yiyuan Liu, Yongqiang Guo, Yuan Ou, Yuduan Wang, Yue Gong, Yuheng Zou, Yujia He, Yunfan Xiong, Yuxiang Luo, Yuxiang You, Yuxuan Liu, Yuyang Zhou, Y~X Zhu, Yanhong Xu, Yanping
  Huang, Yaohui Li, Yi~Zheng, Yuchen Zhu, Yunxian Ma, Ying Tang, Yukun Zha, Yuting Yan, Z~Z Ren, Zehui Ren, Zhangli Sha, Zhe Fu, Zhean Xu, Zhenda Xie, Zhengyan Zhang, Zhewen Hao, Zhicheng Ma, Zhigang Yan, Zhiyu Wu, Zihui Gu, Zijia Zhu, Zijun Liu, Zilin Li, Ziwei Xie, Ziyang Song, Zizheng Pan, Zhen Huang, Zhipeng Xu, Zhongyu Zhang, and Zhen Zhang. 2025.
\newblock {DeepSeek}-{R1}: Incentivizing reasoning capability in {LLMs} via reinforcement learning.
\newblock \emph{arXiv [cs.CL]}.

\bibitem[{Guan et~al.(2025)Guan, Zeng, Meng, Xin, Lu, Lin, Han, Sun, and Zhou}]{guan2025deepragthinkingretrievalstep}
Xinyan Guan, Jiali Zeng, Fandong Meng, Chunlei Xin, Yaojie Lu, Hongyu Lin, Xianpei Han, Le~Sun, and Jie Zhou. 2025.
\newblock \href {http://arxiv.org/abs/2502.01142} {Deeprag: Thinking to retrieval step by step for large language models}.

\bibitem[{Ho et~al.(2020)Ho, Duong~Nguyen, Sugawara, and Aizawa}]{ho-etal-2020-constructing}
Xanh Ho, Anh-Khoa Duong~Nguyen, Saku Sugawara, and Akiko Aizawa. 2020.
\newblock \href {https://doi.org/10.18653/v1/2020.coling-main.580} {Constructing a multi-hop {QA} dataset for comprehensive evaluation of reasoning steps}.
\newblock In \emph{Proceedings of the 28th International Conference on Computational Linguistics}, pages 6609--6625, Barcelona, Spain (Online). International Committee on Computational Linguistics.

\bibitem[{Jiang et~al.(2023)Jiang, Xu, Gao, Sun, Liu, Dwivedi-Yu, Yang, Callan, and Neubig}]{Jiang2023-mk}
Zhengbao Jiang, Frank~F Xu, Luyu Gao, Zhiqing Sun, Qian Liu, Jane Dwivedi-Yu, Yiming Yang, Jamie Callan, and Graham Neubig. 2023.
\newblock Active retrieval augmented generation.
\newblock \emph{arXiv [cs.CL]}.

\bibitem[{Jin et~al.(2025{\natexlab{a}})Jin, Zeng, Yue, Wang, Zamani, and Han}]{Jin2025-ev}
Bowen Jin, Hansi Zeng, Zhenrui Yue, Dong Wang, Hamed Zamani, and Jiawei Han. 2025{\natexlab{a}}.
\newblock Search-{R1}: Training {LLMs} to reason and leverage search engines with reinforcement learning.
\newblock \emph{arXiv [cs.CL]}.

\bibitem[{Jin et~al.(2025{\natexlab{b}})Jin, Zeng, Yue, Yoon, Arik, Wang, Zamani, and Han}]{jin2025searchr1trainingllmsreason}
Bowen Jin, Hansi Zeng, Zhenrui Yue, Jinsung Yoon, Sercan Arik, Dong Wang, Hamed Zamani, and Jiawei Han. 2025{\natexlab{b}}.
\newblock \href {http://arxiv.org/abs/2503.09516} {Search-r1: Training llms to reason and leverage search engines with reinforcement learning}.

\bibitem[{Joshi et~al.(2017)Joshi, Choi, Weld, and Zettlemoyer}]{Joshi2017-qg}
Mandar Joshi, Eunsol Choi, Daniel~S Weld, and Luke Zettlemoyer. 2017.
\newblock {TriviaQA}: A large scale distantly supervised challenge dataset for reading comprehension.

\bibitem[{Karpukhin et~al.(2020)Karpukhin, Oğuz, Min, Lewis, Wu, Edunov, Chen, and Yih}]{Karpukhin2020-tv}
Vladimir Karpukhin, Barlas Oğuz, Sewon Min, Patrick Lewis, Ledell Wu, Sergey Edunov, Danqi Chen, and Wen-Tau Yih. 2020.
\newblock Dense passage retrieval for open-domain question answering.
\newblock \emph{arXiv [cs.CL]}.

\bibitem[{Kwiatkowski et~al.(2019)Kwiatkowski, Palomaki, Redfield, Collins, Parikh, Alberti, Epstein, Polosukhin, Devlin, Lee, Toutanova, Jones, Kelcey, Chang, Dai, Uszkoreit, Le, and Petrov}]{Kwiatkowski2019-zk}
Tom Kwiatkowski, Jennimaria Palomaki, Olivia Redfield, Michael Collins, Ankur Parikh, Chris Alberti, Danielle Epstein, Illia Polosukhin, Jacob Devlin, Kenton Lee, Kristina Toutanova, Llion Jones, Matthew Kelcey, Ming-Wei Chang, Andrew~M Dai, Jakob Uszkoreit, Quoc Le, and Slav Petrov. 2019.
\newblock Natural questions: A benchmark for question answering research.
\newblock \emph{Trans. Assoc. Comput. Linguist.}, 7:453--466.

\bibitem[{Lewis et~al.(2020)Lewis, Perez, Piktus, Petroni, Karpukhin, Goyal, Küttler, Lewis, Yih, Rocktäschel, Riedel, and Kiela}]{Lewis2020-ik}
Patrick Lewis, Ethan Perez, Aleksandra Piktus, Fabio Petroni, Vladimir Karpukhin, Naman Goyal, Heinrich Küttler, Mike Lewis, Wen-Tau Yih, Tim Rocktäschel, Sebastian Riedel, and Douwe Kiela. 2020.
\newblock Retrieval-augmented generation for knowledge-intensive {NLP} tasks.
\newblock \emph{arXiv [cs.CL]}.

\bibitem[{Li et~al.(2025)Li, Dong, Jin, Zhang, Zhou, Zhu, Zhang, and Dou}]{Li2025-ws}
Xiaoxi Li, Guanting Dong, Jiajie Jin, Yuyao Zhang, Yujia Zhou, Yutao Zhu, Peitian Zhang, and Zhicheng Dou. 2025.
\newblock Search-{o1}: Agentic search-enhanced large reasoning models.
\newblock \emph{arXiv [cs.AI]}.

\bibitem[{Mallen et~al.(2023)Mallen, Asai, Zhong, Das, Khashabi, and Hajishirzi}]{Mallen2023-eq}
Alex Mallen, Akari Asai, Victor Zhong, Rajarshi Das, Daniel Khashabi, and Hannaneh Hajishirzi. 2023.
\newblock When not to trust language models: Investigating effectiveness of parametric and non-parametric memories.
\newblock In \emph{Proceedings of the 61st Annual Meeting of the Association for Computational Linguistics (Volume 1: Long Papers)}, Stroudsburg, PA, USA. Association for Computational Linguistics.

\bibitem[{OpenAI et~al.(2024)OpenAI, :, Hurst, Lerer, Goucher, Perelman, Ramesh, Clark, Ostrow, Welihinda, Hayes, Radford, Mądry, Baker-Whitcomb, Beutel, Borzunov, Carney, Chow, Kirillov, Nichol, Paino, Renzin, Passos, Kirillov, Christakis, Conneau, Kamali, Jabri, Moyer, Tam, Crookes, Tootoochian, Tootoonchian, Kumar, Vallone, Karpathy, Braunstein, Cann, Codispoti, Galu, Kondrich, Tulloch, Mishchenko, Baek, Jiang, Pelisse, Woodford, Gosalia, Dhar, Pantuliano, Nayak, Oliver, Zoph, Ghorbani, Leimberger, Rossen, Sokolowsky, Wang, Zweig, Hoover, Samic, McGrew, Spero, Giertler, Cheng, Lightcap, Walkin, Quinn, Guarraci, Hsu, Kellogg, Eastman, Lugaresi, Wainwright, Bassin, Hudson, Chu, Nelson, Li, Shern, Conger, Barette, Voss, Ding, Lu, Zhang, Beaumont, Hallacy, Koch, Gibson, Kim, Choi, McLeavey, Hesse, Fischer, Winter, Czarnecki, Jarvis, Wei, Koumouzelis, Sherburn, Kappler, Levin, Levy, Carr, Farhi, Mely, Robinson, Sasaki, Jin, Valladares, Tsipras, Li, Nguyen, Findlay, Oiwoh, Wong, Asdar, Proehl, Yang, Antonow,
  Kramer, Peterson, Sigler, Wallace, Brevdo, Mays, Khorasani, Such, Raso, Zhang, von Lohmann, Sulit, Goh, Oden, Salmon, Starace, Brockman, Salman, Bao, Hu, Wong, Wang, Schmidt, Whitney, Jun, Kirchner, de~Oliveira~Pinto, Ren, Chang, Chung, Kivlichan, O'Connell, O'Connell, Osband, Silber, Sohl, Okuyucu, Lan, Kostrikov, Sutskever, Kanitscheider, Gulrajani, Coxon, Menick, Pachocki, Aung, Betker, Crooks, Lennon, Kiros, Leike, Park, Kwon, Phang, Teplitz, Wei, Wolfe, Chen, Harris, Varavva, Lee, Shieh, Lin, Yu, Weng, Tang, Yu, Jang, Candela, Beutler, Landers, Parish, Heidecke, Schulman, Lachman, McKay, Uesato, Ward, Kim, Huizinga, Sitkin, Kraaijeveld, Gross, Kaplan, Snyder, Achiam, Jiao, Lee, Zhuang, Harriman, Fricke, Hayashi, Singhal, Shi, Karthik, Wood, Rimbach, Hsu, Nguyen, Gu-Lemberg, Button, Liu, Howe, Muthukumar, Luther, Ahmad, Kai, Itow, Workman, Pathak, Chen, Jing, Guy, Fedus, Zhou, Mamitsuka, Weng, McCallum, Held, Ouyang, Feuvrier, Zhang, Kondraciuk, Kaiser, Hewitt, Metz, Doshi, Aflak, Simens, Boyd,
  Thompson, Dukhan, Chen, Gray, Hudnall, Zhang, Aljubeh, Litwin, Zeng, Johnson, Shetty, Gupta, Shah, Yatbaz, Yang, Zhong, Glaese, Chen, Janner, Lampe, Petrov, Wu, Wang, Fradin, Pokrass, Castro, de~Castro, Pavlov, Brundage, Wang, Khan, Murati, Bavarian, Lin, Yesildal, Soto, Gimelshein, Cone, Staudacher, Summers, LaFontaine, Chowdhury, Ryder, Stathas, Turley, Tezak, Felix, Kudige, Keskar, Deutsch, Bundick, Puckett, Nachum, Okelola, Boiko, Murk, Jaffe, Watkins, Godement, Campbell-Moore, Chao, McMillan, Belov, Su, Bak, Bakkum, Deng, Dolan, Hoeschele, Welinder, Tillet, Pronin, Tillet, Dhariwal, Yuan, Dias, Lim, Arora, Troll, Lin, Lopes, Puri, Miyara, Leike, Gaubert, Zamani, Wang, Donnelly, Honsby, Smith, Sahai, Ramchandani, Huet, Carmichael, Zellers, Chen, Chen, Nigmatullin, Cheu, Jain, Altman, Schoenholz, Toizer, Miserendino, Agarwal, Culver, Ethersmith, Gray, Grove, Metzger, Hermani, Jain, Zhao, Wu, Jomoto, Wu, Shuaiqi, Xia, Phene, Papay, Narayanan, Coffey, Lee, Hall, Balaji, Broda, Stramer, Xu, Gogineni,
  Christianson, Sanders, Patwardhan, Cunninghman, Degry, Dimson, Raoux, Shadwell, Zheng, Underwood, Markov, Sherbakov, Rubin, Stasi, Kaftan, Heywood, Peterson, Walters, Eloundou, Qi, Moeller, Monaco, Kuo, Fomenko, Chang, Zheng, Zhou, Manassra, Sheu, Zaremba, Patil, Qian, Kim, Cheng, Zhang, He, Zhang, Jin, Dai, and Malkov}]{openai2024gpt4ocard}
OpenAI, :, Aaron Hurst, Adam Lerer, Adam~P. Goucher, Adam Perelman, Aditya Ramesh, Aidan Clark, AJ~Ostrow, Akila Welihinda, Alan Hayes, Alec Radford, Aleksander Mądry, Alex Baker-Whitcomb, Alex Beutel, Alex Borzunov, Alex Carney, Alex Chow, Alex Kirillov, Alex Nichol, Alex Paino, Alex Renzin, Alex~Tachard Passos, Alexander Kirillov, Alexi Christakis, Alexis Conneau, Ali Kamali, Allan Jabri, Allison Moyer, Allison Tam, Amadou Crookes, Amin Tootoochian, Amin Tootoonchian, Ananya Kumar, Andrea Vallone, Andrej Karpathy, Andrew Braunstein, Andrew Cann, Andrew Codispoti, Andrew Galu, Andrew Kondrich, Andrew Tulloch, Andrey Mishchenko, Angela Baek, Angela Jiang, Antoine Pelisse, Antonia Woodford, Anuj Gosalia, Arka Dhar, Ashley Pantuliano, Avi Nayak, Avital Oliver, Barret Zoph, Behrooz Ghorbani, Ben Leimberger, Ben Rossen, Ben Sokolowsky, Ben Wang, Benjamin Zweig, Beth Hoover, Blake Samic, Bob McGrew, Bobby Spero, Bogo Giertler, Bowen Cheng, Brad Lightcap, Brandon Walkin, Brendan Quinn, Brian Guarraci, Brian Hsu,
  Bright Kellogg, Brydon Eastman, Camillo Lugaresi, Carroll Wainwright, Cary Bassin, Cary Hudson, Casey Chu, Chad Nelson, Chak Li, Chan~Jun Shern, Channing Conger, Charlotte Barette, Chelsea Voss, Chen Ding, Cheng Lu, Chong Zhang, Chris Beaumont, Chris Hallacy, Chris Koch, Christian Gibson, Christina Kim, Christine Choi, Christine McLeavey, Christopher Hesse, Claudia Fischer, Clemens Winter, Coley Czarnecki, Colin Jarvis, Colin Wei, Constantin Koumouzelis, Dane Sherburn, Daniel Kappler, Daniel Levin, Daniel Levy, David Carr, David Farhi, David Mely, David Robinson, David Sasaki, Denny Jin, Dev Valladares, Dimitris Tsipras, Doug Li, Duc~Phong Nguyen, Duncan Findlay, Edede Oiwoh, Edmund Wong, Ehsan Asdar, Elizabeth Proehl, Elizabeth Yang, Eric Antonow, Eric Kramer, Eric Peterson, Eric Sigler, Eric Wallace, Eugene Brevdo, Evan Mays, Farzad Khorasani, Felipe~Petroski Such, Filippo Raso, Francis Zhang, Fred von Lohmann, Freddie Sulit, Gabriel Goh, Gene Oden, Geoff Salmon, Giulio Starace, Greg Brockman, Hadi
  Salman, Haiming Bao, Haitang Hu, Hannah Wong, Haoyu Wang, Heather Schmidt, Heather Whitney, Heewoo Jun, Hendrik Kirchner, Henrique~Ponde de~Oliveira~Pinto, Hongyu Ren, Huiwen Chang, Hyung~Won Chung, Ian Kivlichan, Ian O'Connell, Ian O'Connell, Ian Osband, Ian Silber, Ian Sohl, Ibrahim Okuyucu, Ikai Lan, Ilya Kostrikov, Ilya Sutskever, Ingmar Kanitscheider, Ishaan Gulrajani, Jacob Coxon, Jacob Menick, Jakub Pachocki, James Aung, James Betker, James Crooks, James Lennon, Jamie Kiros, Jan Leike, Jane Park, Jason Kwon, Jason Phang, Jason Teplitz, Jason Wei, Jason Wolfe, Jay Chen, Jeff Harris, Jenia Varavva, Jessica~Gan Lee, Jessica Shieh, Ji~Lin, Jiahui Yu, Jiayi Weng, Jie Tang, Jieqi Yu, Joanne Jang, Joaquin~Quinonero Candela, Joe Beutler, Joe Landers, Joel Parish, Johannes Heidecke, John Schulman, Jonathan Lachman, Jonathan McKay, Jonathan Uesato, Jonathan Ward, Jong~Wook Kim, Joost Huizinga, Jordan Sitkin, Jos Kraaijeveld, Josh Gross, Josh Kaplan, Josh Snyder, Joshua Achiam, Joy Jiao, Joyce Lee, Juntang
  Zhuang, Justyn Harriman, Kai Fricke, Kai Hayashi, Karan Singhal, Katy Shi, Kavin Karthik, Kayla Wood, Kendra Rimbach, Kenny Hsu, Kenny Nguyen, Keren Gu-Lemberg, Kevin Button, Kevin Liu, Kiel Howe, Krithika Muthukumar, Kyle Luther, Lama Ahmad, Larry Kai, Lauren Itow, Lauren Workman, Leher Pathak, Leo Chen, Li~Jing, Lia Guy, Liam Fedus, Liang Zhou, Lien Mamitsuka, Lilian Weng, Lindsay McCallum, Lindsey Held, Long Ouyang, Louis Feuvrier, Lu~Zhang, Lukas Kondraciuk, Lukasz Kaiser, Luke Hewitt, Luke Metz, Lyric Doshi, Mada Aflak, Maddie Simens, Madelaine Boyd, Madeleine Thompson, Marat Dukhan, Mark Chen, Mark Gray, Mark Hudnall, Marvin Zhang, Marwan Aljubeh, Mateusz Litwin, Matthew Zeng, Max Johnson, Maya Shetty, Mayank Gupta, Meghan Shah, Mehmet Yatbaz, Meng~Jia Yang, Mengchao Zhong, Mia Glaese, Mianna Chen, Michael Janner, Michael Lampe, Michael Petrov, Michael Wu, Michele Wang, Michelle Fradin, Michelle Pokrass, Miguel Castro, Miguel Oom~Temudo de~Castro, Mikhail Pavlov, Miles Brundage, Miles Wang, Minal
  Khan, Mira Murati, Mo~Bavarian, Molly Lin, Murat Yesildal, Nacho Soto, Natalia Gimelshein, Natalie Cone, Natalie Staudacher, Natalie Summers, Natan LaFontaine, Neil Chowdhury, Nick Ryder, Nick Stathas, Nick Turley, Nik Tezak, Niko Felix, Nithanth Kudige, Nitish Keskar, Noah Deutsch, Noel Bundick, Nora Puckett, Ofir Nachum, Ola Okelola, Oleg Boiko, Oleg Murk, Oliver Jaffe, Olivia Watkins, Olivier Godement, Owen Campbell-Moore, Patrick Chao, Paul McMillan, Pavel Belov, Peng Su, Peter Bak, Peter Bakkum, Peter Deng, Peter Dolan, Peter Hoeschele, Peter Welinder, Phil Tillet, Philip Pronin, Philippe Tillet, Prafulla Dhariwal, Qiming Yuan, Rachel Dias, Rachel Lim, Rahul Arora, Rajan Troll, Randall Lin, Rapha~Gontijo Lopes, Raul Puri, Reah Miyara, Reimar Leike, Renaud Gaubert, Reza Zamani, Ricky Wang, Rob Donnelly, Rob Honsby, Rocky Smith, Rohan Sahai, Rohit Ramchandani, Romain Huet, Rory Carmichael, Rowan Zellers, Roy Chen, Ruby Chen, Ruslan Nigmatullin, Ryan Cheu, Saachi Jain, Sam Altman, Sam Schoenholz, Sam
  Toizer, Samuel Miserendino, Sandhini Agarwal, Sara Culver, Scott Ethersmith, Scott Gray, Sean Grove, Sean Metzger, Shamez Hermani, Shantanu Jain, Shengjia Zhao, Sherwin Wu, Shino Jomoto, Shirong Wu, Shuaiqi, Xia, Sonia Phene, Spencer Papay, Srinivas Narayanan, Steve Coffey, Steve Lee, Stewart Hall, Suchir Balaji, Tal Broda, Tal Stramer, Tao Xu, Tarun Gogineni, Taya Christianson, Ted Sanders, Tejal Patwardhan, Thomas Cunninghman, Thomas Degry, Thomas Dimson, Thomas Raoux, Thomas Shadwell, Tianhao Zheng, Todd Underwood, Todor Markov, Toki Sherbakov, Tom Rubin, Tom Stasi, Tomer Kaftan, Tristan Heywood, Troy Peterson, Tyce Walters, Tyna Eloundou, Valerie Qi, Veit Moeller, Vinnie Monaco, Vishal Kuo, Vlad Fomenko, Wayne Chang, Weiyi Zheng, Wenda Zhou, Wesam Manassra, Will Sheu, Wojciech Zaremba, Yash Patil, Yilei Qian, Yongjik Kim, Youlong Cheng, Yu~Zhang, Yuchen He, Yuchen Zhang, Yujia Jin, Yunxing Dai, and Yury Malkov. 2024.
\newblock \href {http://arxiv.org/abs/2410.21276} {Gpt-4o system card}.

\bibitem[{Press et~al.(2023)Press, Zhang, Min, Schmidt, Smith, and Lewis}]{press-etal-2023-measuring}
Ofir Press, Muru Zhang, Sewon Min, Ludwig Schmidt, Noah Smith, and Mike Lewis. 2023.
\newblock \href {https://doi.org/10.18653/v1/2023.findings-emnlp.378} {Measuring and narrowing the compositionality gap in language models}.
\newblock In \emph{Findings of the Association for Computational Linguistics: EMNLP 2023}, pages 5687--5711, Singapore. Association for Computational Linguistics.

\bibitem[{Qian et~al.(2025)Qian, Acikgoz, Wang, Chen, Sil, Hakkani-Tür, Tur, and Ji}]{qian2025smartselfawareagenttool}
Cheng Qian, Emre~Can Acikgoz, Hongru Wang, Xiusi Chen, Avirup Sil, Dilek Hakkani-Tür, Gokhan Tur, and Heng Ji. 2025.
\newblock \href {http://arxiv.org/abs/2502.11435} {Smart: Self-aware agent for tool overuse mitigation}.

\bibitem[{Qwen et~al.(2025)Qwen, :, Yang, Yang, Zhang, Hui, Zheng, Yu, Li, Liu, Huang, Wei, Lin, Yang, Tu, Zhang, Yang, Yang, Zhou, Lin, Dang, Lu, Bao, Yang, Yu, Li, Xue, Zhang, Zhu, Men, Lin, Li, Tang, Xia, Ren, Ren, Fan, Su, Zhang, Wan, Liu, Cui, Zhang, and Qiu}]{qwen2025qwen25technicalreport}
Qwen, :, An~Yang, Baosong Yang, Beichen Zhang, Binyuan Hui, Bo~Zheng, Bowen Yu, Chengyuan Li, Dayiheng Liu, Fei Huang, Haoran Wei, Huan Lin, Jian Yang, Jianhong Tu, Jianwei Zhang, Jianxin Yang, Jiaxi Yang, Jingren Zhou, Junyang Lin, Kai Dang, Keming Lu, Keqin Bao, Kexin Yang, Le~Yu, Mei Li, Mingfeng Xue, Pei Zhang, Qin Zhu, Rui Men, Runji Lin, Tianhao Li, Tianyi Tang, Tingyu Xia, Xingzhang Ren, Xuancheng Ren, Yang Fan, Yang Su, Yichang Zhang, Yu~Wan, Yuqiong Liu, Zeyu Cui, Zhenru Zhang, and Zihan Qiu. 2025.
\newblock \href {http://arxiv.org/abs/2412.15115} {Qwen2.5 technical report}.

\bibitem[{Shao et~al.(2024)Shao, Wang, Zhu, Xu, Song, Zhang, Li, Wu, and Guo}]{Shao2024-dp}
Zhihong Shao, Peiyi Wang, Qihao Zhu, Runxin Xu, Junxiao Song, Mingchuan Zhang, Y~K Li, Y~Wu, and Daya Guo. 2024.
\newblock {DeepSeekMath}: Pushing the limits of mathematical reasoning in open language models.
\newblock \emph{arXiv [cs.CL]}.

\bibitem[{Shen et~al.(2024)Shen, Zhu, and Chen}]{shen-etal-2024-smartcal}
Yuanhao Shen, Xiaodan Zhu, and Lei Chen. 2024.
\newblock \href {https://doi.org/10.18653/v1/2024.emnlp-industry.59} {{SMARTCAL}: An approach to self-aware tool-use evaluation and calibration}.
\newblock In \emph{Proceedings of the 2024 Conference on Empirical Methods in Natural Language Processing: Industry Track}, pages 774--789, Miami, Florida, US. Association for Computational Linguistics.

\bibitem[{Song et~al.(2025{\natexlab{a}})Song, Jiang, Min, Chen, Chen, Zhao, Fang, and Wen}]{Song2025-hj}
Huatong Song, Jinhao Jiang, Yingqian Min, Jie Chen, Zhipeng Chen, Wayne~Xin Zhao, Lei Fang, and Ji-Rong Wen. 2025{\natexlab{a}}.
\newblock {R1}-searcher: Incentivizing the search capability in {LLMs} via reinforcement learning.
\newblock \emph{arXiv [cs.AI]}.

\bibitem[{Song et~al.(2025{\natexlab{b}})Song, Jiang, Min, Chen, Chen, Zhao, Fang, and Wen}]{song2025r1searcherincentivizingsearchcapability}
Huatong Song, Jinhao Jiang, Yingqian Min, Jie Chen, Zhipeng Chen, Wayne~Xin Zhao, Lei Fang, and Ji-Rong Wen. 2025{\natexlab{b}}.
\newblock \href {http://arxiv.org/abs/2503.05592} {R1-searcher: Incentivizing the search capability in llms via reinforcement learning}.

\bibitem[{Trivedi et~al.(2022)Trivedi, Balasubramanian, Khot, and Sabharwal}]{trivedi-etal-2022-musique}
Harsh Trivedi, Niranjan Balasubramanian, Tushar Khot, and Ashish Sabharwal. 2022.
\newblock \href {https://doi.org/10.1162/tacl_a_00475} {{M}u{S}i{Q}ue: Multihop questions via single-hop question composition}.
\newblock \emph{Transactions of the Association for Computational Linguistics}, 10:539--554.

\bibitem[{Trivedi et~al.(2023)Trivedi, Balasubramanian, Khot, and Sabharwal}]{Trivedi2023-cl}
Harsh Trivedi, Niranjan Balasubramanian, Tushar Khot, and Ashish Sabharwal. 2023.
\newblock Interleaving retrieval with chain-of-thought reasoning for knowledge-intensive multi-step questions.
\newblock In \emph{Proceedings of the 61st Annual Meeting of the Association for Computational Linguistics (Volume 1: Long Papers)}, Stroudsburg, PA, USA. Association for Computational Linguistics.

\bibitem[{Wang et~al.(2025{\natexlab{a}})Wang, Qian, Zhong, Chen, Qiu, Huang, Jin, Wang, Wong, and Ji}]{wang2025otcoptimaltoolcalls}
Hongru Wang, Cheng Qian, Wanjun Zhong, Xiusi Chen, Jiahao Qiu, Shijue Huang, Bowen Jin, Mengdi Wang, Kam-Fai Wong, and Heng Ji. 2025{\natexlab{a}}.
\newblock \href {http://arxiv.org/abs/2504.14870} {Otc: Optimal tool calls via reinforcement learning}.

\bibitem[{Wang et~al.(2025{\natexlab{b}})Wang, Chen, Yang, Huang, Dou, and Wei}]{wang2025chainofretrievalaugmentedgeneration}
Liang Wang, Haonan Chen, Nan Yang, Xiaolong Huang, Zhicheng Dou, and Furu Wei. 2025{\natexlab{b}}.
\newblock \href {http://arxiv.org/abs/2501.14342} {Chain-of-retrieval augmented generation}.

\bibitem[{Wang et~al.(2022)Wang, Yang, Huang, Jiao, Yang, Jiang, Majumder, and Wei}]{Wang2022-rb}
Liang Wang, Nan Yang, Xiaolong Huang, Binxing Jiao, Linjun Yang, Daxin Jiang, Rangan Majumder, and Furu Wei. 2022.
\newblock Text embeddings by weakly-supervised contrastive pre-training.
\newblock \emph{arXiv [cs.CL]}.

\bibitem[{Wei et~al.(2022)Wei, Wang, Schuurmans, Bosma, Ichter, Xia, Chi, Le, and Zhou}]{Wei2022-ds}
Jason Wei, Xuezhi Wang, Dale Schuurmans, Maarten Bosma, Brian Ichter, Fei Xia, Ed~Chi, Quoc Le, and Denny Zhou. 2022.
\newblock Chain-of-thought prompting elicits reasoning in large language models.
\newblock \emph{arXiv [cs.CL]}.

\bibitem[{Yang et~al.(2018)Yang, Qi, Zhang, Bengio, Cohen, Salakhutdinov, and Manning}]{yang-etal-2018-hotpotqa}
Zhilin Yang, Peng Qi, Saizheng Zhang, Yoshua Bengio, William Cohen, Ruslan Salakhutdinov, and Christopher~D. Manning. 2018.
\newblock \href {https://doi.org/10.18653/v1/D18-1259} {{H}otpot{QA}: A dataset for diverse, explainable multi-hop question answering}.
\newblock In \emph{Proceedings of the 2018 Conference on Empirical Methods in Natural Language Processing}, pages 2369--2380, Brussels, Belgium. Association for Computational Linguistics.

\end{thebibliography}

% Appendix comes after the references and must start at a new page
\cleardoublepage\appendix
\section{Appendix}
\label{sec:appendix}

\subsection{Detailed Step-wise Analysis Procedure}
\label{ssec:detailed-step-wise-analysis-procedure}
To empirically measure the rates of over-search and under-search, we conducted a detailed step-wise analysis of the agent's decision-making process. The interactions of the agent are logged as a sequence of steps, where each step can involve internal reasoning (thinking), querying a search tool, processing retrieved context, and generating an answer. We define specific procedures to identify and quantify each type of sub-optimal search behavior:
\begin{enumerate}
    \item \textbf{Step Extraction:} We parse the agent's interaction log following the definition in Appendix \ref{ssec:formal-definition-of-over-search-under-search}. Each distinct thinking process is a decision point and considered a step, typically delineated by <step> and </step> tags (or a similar structured logging format). A "search step" is identified as any step where all three relevant operations—think (the model's reasoning), search (the search query issued), context (the information retrieved). A "non-search step" typically only consists of thinking. In this work specifically, the Step Extraction is done by prompting QwQ-32B \cite{qwen2025qwen25technicalreport} as we discover that reasoning LLM typically perform better on such task. An example of instruction for step extraction is provided in Appendix \ref{ssec:instruction_step_extraction}.

    \item \textbf{Extraction of Partial Input:} For each identified search step, we reconstruct the input that would have been available to the model before it decided to search. This is achieved by taking the complete output generated by the agent from the beginning of the interaction up to and including the content of the think field of the current search step.

    \item \textbf{Querying with Internal Knowledge for Over-search Analysis:} For over-search rate measurement, the extracted partial output is then appended with a specific instructional prompt: "I will use my own knowledge to answer this query and provide my answer to this query enclosed in <query\_answer> </query\_answer> tags." This combined text serves as a new input to the original RL-tuned model (e.g., Search-R1-$\beta$-GRPO and Search-R1-GRPO), which is tasked with generating an answer without performing any new search. The over-search rate is then measured by computing the percentage of steps that provide equivalent answer (determined by QwQ-32B in our analysis) for both with and without searching, among all "search steps".

    \item \textbf{Generation of Reference Answer for Under-search Analysis:} For each identified non-search step, the original query or sub-query that the agent was attempting to answer at that point is presented to a more powerful, state-of-the-art language model (e.g., ChatGPT-4o \cite{openai2024gpt4ocard}) with recent knowledge cutoff date. This model generates a "reference answer," which is assumed to be of high quality. The reference answer obtained is compared with the actual answer generated by the agent for that non-search step. The under-search rate is calculated as the proportion of non-search steps where the agent's answer does not match (determined by QwQ-32B in our analysis) the reference answer, quantifying how often the agent fails to search when doing so would have likely led to a more accurate or complete answer.

\end{enumerate}

\subsection{Search Frequency vs. Optimal Hops}
\label{ssec:search-frequency-vs-optimal-hops}
One indicator of potential over-search is when the number of search queries generated by an agent exceeds the optimal number of reasoning hops required to answer a question. A significantly higher search count often points to redundant information gathering. For this experiment, we only use the test set from Bamboogle \cite{press-etal-2023-measuring} and MuSiQue \cite{trivedi-etal-2022-musique} as they are the only two datasets providing pre-defined number of hops for each test sample.

\begin{table}[t!]
\centering\resizebox{\columnwidth}{!}{
\begin{tabular}{ll|cccc}
\toprule
\bf Model       & \bf Dataset   & \bf Search vs. Hops & \bf Correct (\%) & \bf Incorrect (\%) & \bf Sum (\%) \\
\midrule
 &   & Less            & 2.8         & 19            & 21.8    \\
R1-Searcher  & Musique   & Match           & 21.8        & 45.8          & 67.6    \\
             &           & More            & 1.8         & 8.8           & 10.6    \\
\midrule
 &  & Less            & 0           & 0             & 0       \\
R1-Searcher  & Bamboogle & Match           & 40.8        & 52.8          & 93.6    \\
             &           & More            & 3.2         & 3.2           & 6.4     \\
\midrule
   &    & Less            & 1.8         & 7             & 8.8     \\
Search-R1    & Musique & Match           & 12.4        & 27.6          & 40      \\
             &           & More            & 8.8         & 42.4          & 51.2    \\
\midrule
   &  & Less            & 0.8         & 1.6           & 2.4     \\
Search-R1    & Bamboogle & Match           & 28.8        & 28            & 56.8    \\
             &           & More            & 12          & 28.8          & 40.8    \\
\bottomrule
\end{tabular}}
\caption{Comparison of the number of searches generated vs. annotated hops on Bamboogle and Musique datasets. "More" indicates potential over-search as number of searchers exceeds pre-defined optimal hops. "Less" may indicate a potential under-search.}
% \vspace{-0.2in}
\label{tab:search_vs_hops}
\end{table}

R1-Searcher exhibits a tendency to perform more searches than hops in 10.6\% of Musique cases and 6.4\% of Bamboogle cases. Search-R1 shows a more pronounced tendency, with 51.2\% (Musique) and 40.8\% (Bamboogle) of cases issuing more searches than annotated hops. This result suggests that models trained with different methods do not inherently solve over-search and might even exacerbate it under certain configurations if not properly guided. While "Less" searches than hops might indicate efficient reasoning or under-search, the "More" category strongly suggests instances of over-searching.

\subsection{Instruction for Model Input}\label{apx:instruction}
\begin{AIbox}
Answer the given question.  You must conduct reasoning inside <think> and </think> first every time you get new information.  After reasoning, if you find you lack some knowledge, you can call a search engine by <search> query </search>, and it will return the top searched results between <information> and </information>.   You can search as many times as you want.  If you find no further external knowledge needed, you can directly provide the answer inside <answer> and </answer> without detailed illustrations. For example, <answer> Beijing </answer>. Question: question.
\end{AIbox}

\subsection{Instruction for Step Extraction}
\label{ssec:instruction_step_extraction}
\begin{framed}
\textbf{Objective:} Your task is to parse a complete interaction log from a reasoning agent. Your goal is to segment the log into a chronological sequence of steps and structure the output for each step into a consistent JSON format.
\vspace{\baselineskip}

\textbf{Core Methodology:} A reasoning trajectory is a sequence of steps, $s_1, s_2, \dots, s_N$. Each step, $s_t$, involves a reasoning component, $r_t$. The sub-answer, $a_t$, which is the conclusion of step $s_t$, is reflected in the reasoning component of the \textbf{next} step, $r_{t+1}$. You must follow this look-ahead method to determine the conclusion for each step.
\vspace{\baselineskip}

\textbf{Key Definitions:}
\begin{description}
    \item[Search Step:] A step where the agent uses a search tool. It must include a reasoning block (\texttt{<think>}), a search query (\texttt{<search>}), and retrieved context (\texttt{<information>}).
    \item[Non-Search Step:] A step where the agent relies only on its internal knowledge and prior context. It typically only includes a reasoning block (\texttt{<think>}).
    \item[Conclusion:] The sub-answer or piece of information the agent generates or confirms at the end of a step ($s_t$), which contributes to the final answer.
\end{description}
\vspace{\baselineskip}

\textbf{Instructions:}
\begin{enumerate}[label=\arabic*., nosep]
    \item Parse the entire interaction log into a chronological sequence of steps. A new step begins with each distinct reasoning block (e.g., content within \texttt{<think>} tags).

    \item For each step ($s_t$), extract the following components:
    \begin{itemize}[nosep]
        \item \textbf{reasoning:} The content from the \texttt{<think>} block of the current step.
        \item \textbf{query:} The content from the \texttt{<search>} block of the current step. If not present, use \texttt{null}.
        \item \textbf{information:} The content from the \texttt{<information>} block of the current step. If not present, use \texttt{null}.
        \item \textbf{conclusion:} The specific sub-answer ($a_t$) produced as a result of the current step's actions. \textbf{Crucially, you must identify this conclusion by analyzing the reasoning block of the \textit{following} step ($r_{t+1}$), where it is first used or stated.} For the final step in the trajectory, the conclusion is the final answer itself.
    \end{itemize}

    \item For each step, construct a JSON object containing the extracted components.

    \item Present the final output as a sequence of \texttt{<step>} blocks, with each block containing the JSON object for that step.
\end{enumerate}
\vspace{\baselineskip}

\textbf{Required Output Format:} Your entire output must be a sequence of \texttt{<step>} blocks.

\textbf{Example for a Search Step:}
\begin{lstlisting}[language=XML, style=jsonstyle, basicstyle=\ttfamily\footnotesize]
<step>
{
  "reasoning": "The user is asking for the capital of France. I should search for this information to be certain.",
  "query": "capital of France",
  "information": "Paris is the capital and most populous city of France...",
  "conclusion": "The capital of France is Paris."
}
</step>
\end{lstlisting}

\textbf{Example for a Non-Search Step:}
\begin{lstlisting}[language=XML, style=jsonstyle, basicstyle=\ttfamily\footnotesize]
<step>
{
  "reasoning": "Now that I know the capital is Paris, I can formulate the final answer.",
  "query": null,
  "information": null,
  "conclusion": "The final answer is Paris."
}
</step>
\end{lstlisting}
\end{framed}

\subsection{Training Configuration \& Rewards}\label{apx:train_config}
We train Search-R1-GPRO and Search-R1-$\beta$-GPRO for 200 steps, with a learning rate of 1e-6 and batch size of 512. For a question, we produce 5 generations with temperature of 1 to form a GPRO group. For the search engine, for fair comparison, we also use 2018 Wikipedia dump~\cite{Karpukhin2020-tv} as the knowledge source and E5~\cite{Wang2022-rb} as the retriever as Search-R1 and for each search query, top-3 documents are returned. Our training are conducted on two A100 GPUs.

\label{ssec:training_rewards}
\begin{figure}[t]
    \centering
    \includegraphics[width=0.9\columnwidth]{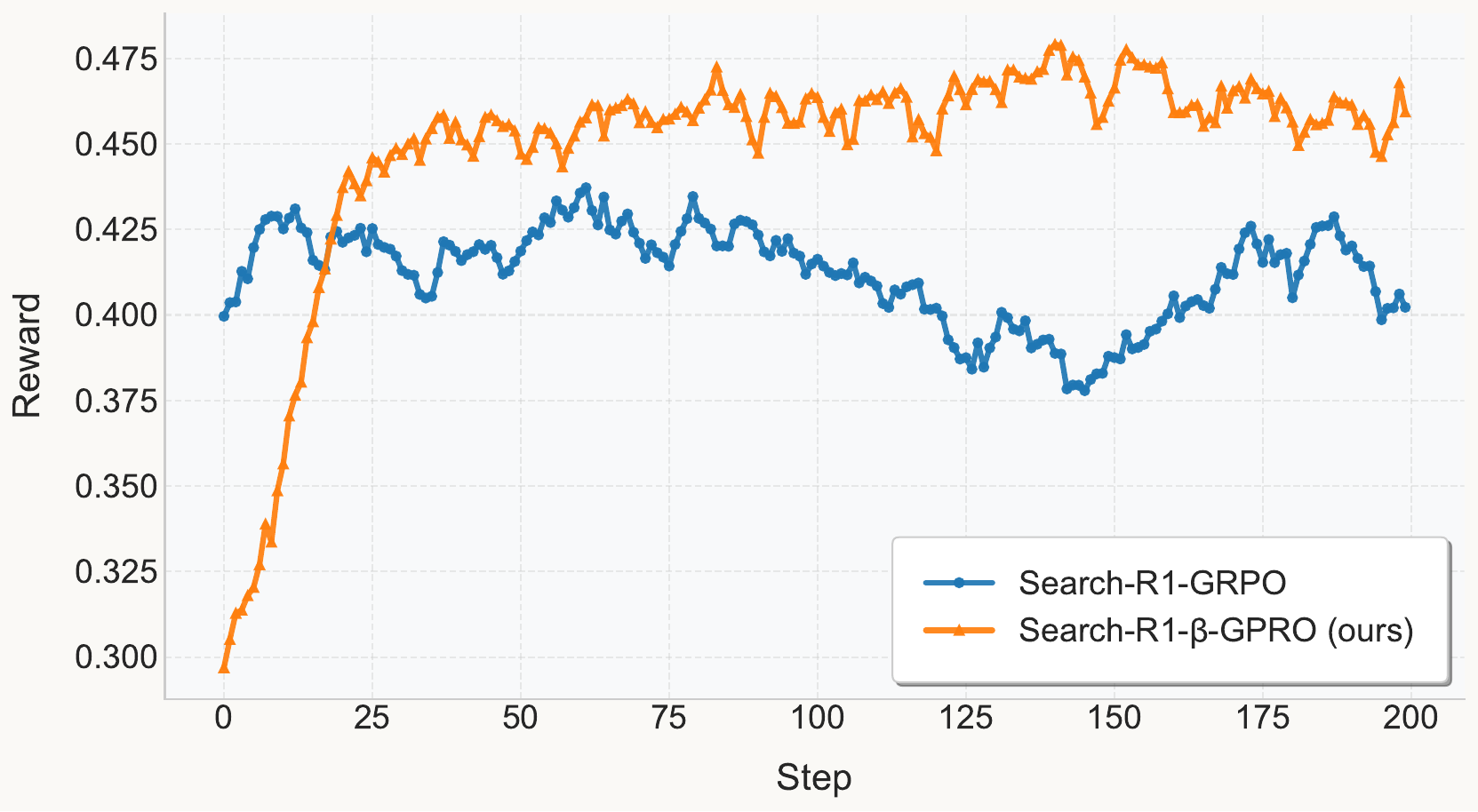}
    \caption{Training Rewards for Search-R1-GRPO and Search-R1-$\beta$-GRPO.}
    \label{fig:reward_curve}
\end{figure}

% \subsection{More Case Study on Sub-optimal Searches}
% \label{ssec:more-case-study=on-sub-optimal-searches}

\end{document}